# An Error Correction Mid-term Electricity Load Forecasting Model Based on Seasonal Decomposition

Liping Zhang
School of Computer Science and Technology,
Chongqing University of Posts and
Telecommunications
Chongqing 400065, China
Chongqing Institute of Green and Intelligent
Technology, Chinese Academy of Sciences
Chongqing School, University of Chinese
Academy of Sciences
Chongqing 400714, China
is.lpzhang@gmail.com

Di Wu, *Member, IEEE*
College of Computer and Information Science,
Southwest University
Chongqing 400715, China
wudi.cigit@gmail.com

Xin Luo, *Senior Member, IEEE*
College of Computer and Information Science,
Southwest University
Chongqing 400715, China
luoxin@swu.edu.cn

*Abstract*—Mid-term electricity load forecasting (LF) plays a critical role in power system planning and operation. To address the issue of error accumulation and transfer during the operation of existing LF models, a novel model called error correction based LF (ECLF) is proposed in this paper, which is designed to provide more accurate and stable LF. Firstly, time series analysis and feature engineering act on the original data to decompose load data into three components and extract relevant features. Then, based on the idea of stacking ensemble, long short-term memory is employed as an error correction module to forecast the components separately, and the forecast results are treated as new features to be fed into extreme gradient boosting for the second-step forecasting. Finally, the component sub-series forecast results are reconstructed to obtain the final LF results. The proposed model is evaluated on real-world electricity load data from two cities in China, and the experimental results demonstrate its superior performance compared to the other benchmark models.

*Keywords—electricity load forecasting, time series analysis, long short-term memory, extreme gradient boosting, error correction, stacking.*

## I. INTRODUCTION

Electricity load forecasting (LF) is a crucial task in the power industry, as it is essential for ensuring reliable and efficient operation of power systems [1]. Accurate LF helps grid operators to make informed decisions regarding generation, transmission, and distribution of electricity which can result in a reliable and efficient power supply to consumers and improved energy management, environmental benefits, and cost savings [2, 3]. Furthermore, it also guides policymakers to make plans for energy policies and infrastructure investments. Given the importance of LF, there is an ongoing need to develop more accurate and reliable forecasting methods [4]. However, accurate LF is a challenging task due to the complex and dynamic nature of electricity demand, which is influenced by various factors such as weather, economic conditions, and consumer behavior [5-7].

There has been significant research in the field of mid-term LF over the past few decades, which can be roughly divided into two categories: classical time series methods and machine learning (ML) methods. Early studies focused on the former which relied on empirical analysis and linear models based on mathematical and statistical theories. The most commonly used classical methods include regression analysis, autoregressive integrated moving average model (ARIMA), and triple exponential smoothing (TES), which have strong interpretability and are easy to implement [8]. However, due to their relatively simple model structures, they are limited in the ability to comprehensively consider the multiple factors that affect load demand, leading to limited fitting ability [9, 10]. With the emergence of ML, there has been a growing interest in applying it to LF. ML methods [10] based on support vector machines such as support vector regression (SVR), tree models such as extreme gradient boosting (XGBoost) and random forest (RF), and artificial neural networks such as long short-term memory (LSTM) and convolutional neural network (CNN), have been widely used for LF. They can capture both the intrinsic and extrinsic influencing factors and can handle various types and granularities of LF tasks more flexibly than classical methods [10, 11]. However, the weak interpretability and low generalization of pure ML models may lead to unstable and inaccurate predictions that are heavily sensitive to specific parameter combinations and datasets [12, 13].

Furthermore, electricity load data is essentially a time series that exhibits non-linearity and non-stationarity. In such cases, hybrid ML models that combine multiple techniques have become the mainstream method for LF. Time series decomposition can be used to analyze and separate the various complex factors from time series data, providing high-quality data for LF. Hybrid ML models that incorporate time series decomposition have been developed to improve forecast accuracy. For example, to improve robustness and accuracy of LF, a hybrid empirical mode decomposition and state space model was proposed in [14], while [15] incorporated the economic factors into X-12-ARIMA and [16] proposed a characteristic load decomposition and LSTM based forecasting scheme for a mixed-use complex. In addition to exploring the inherent characteristics of the time series itself, it is also important to consider external related information [17-19]. To this end, [20] decomposed the original data and used LSTM to forecast each component with features selected by maximal information coefficient, [21] decomposed the data of load and its features into the same number of sub-series and input them into back propagation network. Various hybrid models have been developed that incorporate external factors to present the complex nonlinear relationships with electricity load demand.

However, the performance of these data-driven models heavily relies on the quality of the training data and complex parameters [22-24], which may lead to error accumulation and propagation during the training process. This can result in significant bias in LF, ultimately limiting their reliability and applicability in practical engineering applications. This is a problem common to current single ML models as well as hybrid ML models. Therefore, this study envisions whether a new model can be devised to eliminate this effect as much as possible, thus achieving more accurate LF to a greater extent. Based on this objective, a new model called error correction based LF (ECLF) is proposed in this paper. Specifically, ECLF decouples the electricity load data via seasonal and trend decomposition using LOESS (STL), meanwhile, selects features for subsequent modeling through feature engineering based on the Pearson correlation coefficient (PCC) algorithm, then adopts ML methods to construct error correction based forecasting modules for sub-series, and finally outputs error-compensated LF by merging sub-series forecast results.

The main contributions of this paper are as follows. Firstly, a new LF model called ECLF is proposed, which effectively integrates the advantages of time series decomposition and ML to achieve more accurate LF. Secondly, based on the idea of stacking ensemble, the error correction based forecasting modules are constructed to forecast the component series, which are finally reconstructed into a complete time series to obtain the optimal output of LF. Finally, the comparison results on real city electricity load consumption data demonstrate the superiority of the proposed forecasting model.

## II. LOAD FORECASTING MODELING

### A. The Proposed Model Structure

Fig. 1 shows the structure of the proposed ECLF model. The forecasting process of ECLF mainly consists of the following four parts.

*a) Load Dataset:* To load the original datasets which contain a set of time series of electricity load consumption $Y$ and its related influencing factors $X$.

*b) Data Processing and Feature Selection:* STL is used for decoupling $Y$ into three components, i.e. seasonal, trend and random sub-series, while PCC analysis and feature importance test are used for extracting features $F_T$ and $F_R$ from $X$ which are more relevant to trend and random components, respectively.

*c) Error Correction based Forecasting:* Seasonal, trend and random sub-series $S$, $T$ and $R$ are obtained after the decompostion of $Y$. For sub-series $T$, $R$ and their feature spaces $F_T$ and $F_R$, LSTM is applied to obtain $T'$ and $R'$, which are supposed to be extended to the original feature spaces to obtain new feature spaces $F_T'$ and $F_R'$. Then, XGBoost is applied to obtain the final forecast results $\hat{T}$ and $\hat{R}$ based on their new feature spaces. For $S$, the model simply takes the values of the same period last year as $\hat{S}$.

*d) Time Series Reconstruction:* Once the three sub-series forecast results $\hat{S}$, $\hat{T}$ and $\hat{R}$ are acquired, they will be reconstructed into one complete time series $\hat{Y}$ (i.e. the LF results) based on STL addictive rule.

### B. Data Processing and Feature Selection

Electricity load demand is essentially a time series with obvious periodicity and uncertainty, which changes mainly with the change of external factors. However, the changes due to these factors may show a phenomenon of masking the basic change trend of load demand on the curve. To analyze the overall trend of load demand, the influence of cyclical and stochastic factors must be separated in order to obtain a curve representing the long-term trend. Considering that the datasets are collected from the load consumption of the whole cities, which has obvious trend and seasonality, the STL algorithm is used to decouple the original electricity load data into three sub-series. STL is a widely-used time series decomposition method which is able to adapt to different seasonal patterns of data and shows good robustness to outliers. The method is based on locally weighted regression (LOESS) and uses a robust iterative scheme to estimate the components of the time series. The seasonal component is estimated by averaging over the seasonal sub-series, the trend component is estimated by fitting a LOESS curve to the data, and the random component is obtained by subtracting the above two components from the original time series. After the decomposition of electricity load data, PCC algorithm and feature importance test are used to pick out features most corresponding to different sub-series. PCC is a statistical index that measures the linear correlation between two continuous variables. Suppose $X = (X_1, X_2 ..., X_n)$ and $Y = (Y_1, Y_2 ..., Y_n)$ are data samples of two consecutive variables, PCC is defined as follows:

$$\rho_{X,Y} = \frac{\sum_{i=1}^{n}(X_i - \bar{X})(Y_i - \bar{Y})}{\sqrt{\sum_{i=1}^{n}(X_i - \bar{X})^2}\sqrt{\sum_{i=1}^{n}(Y_i - \bar{Y})^2}} \qquad (1)$$

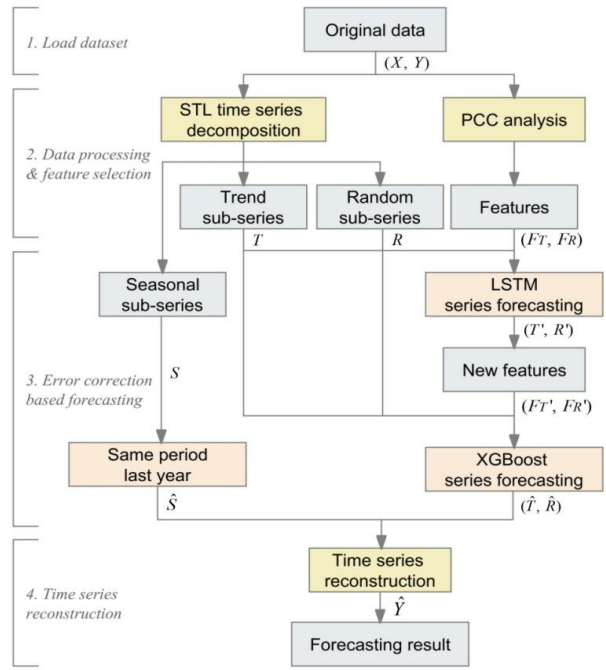

Fig. 1. The proposed model structure.

where $X_i$ and $Y_i$ are the $i$ th sample values of $X$ and $Y$, $\bar{X}$ and $\bar{Y}$ are the mean values of $X$ and $Y$, respectively. The value of PCC ranges from [-1, 1], where 1 indicates complete positive correlation, -1 complete negative correlation, and 0 no correlation. However, the relevant features selected by PCC may lead to redundancy, leading to a degraded forecasting performance. In this paper, the model adopts the feature importance test method according to [9] to remove redundancy features.

*C. Error Correction based Forecasting*

Time series decomposition methods do not allow direct LF, but require subsequent modeling of the sub-series to complete the LF task. The trend component changes slowly, and its regularity depends on the slow-changing factors. The random component is affected by a variety of uncertainty factors, and the regularity of the seasonal component depends on the periodicity of social life and production, which means it tends to show a nearly fixed regularity within each cycle of a time series. To this end, ECLF simply takes the values of the same period last year for seasonal component forecasting, and constructs error correction based forecasting modules for more accurate sub-series forecasting in consideration of the characteristics of trend and random components.

LSTM is a variant of recurrent neural network (RNN) that is good at handling time series forecasting. Compared to traditional RNN, LSTM has improved memory and forgetting mechanisms which effectively handle long-term dependencies. An LSTM cell consists of three gates and a memory cell. The input, forget, and output gates are responsible for controlling the flow of information into and out of the cell, while the memory cell stores information for later use. The equations for the LSTM cell are as follows:

$$i_t = \sigma(W_{xi}x_t + W_{hi}h_{t-1} + b_i) \tag{2}$$

$$f_t = \sigma(W_{xf}x_t + W_{hf}h_{t-1} + b_f) \tag{3}$$

$$o_t = \sigma(W_{xo}x_t + W_{ho}h_{t-1} + b_o) \tag{4}$$

$$\tilde{C}_t = \tanh(W_{xc}x_t + W_{hc}h_{t-1} + b_c) \tag{5}$$

$$C_t = f_t \odot C_{t-1} + i_t \odot \tilde{C}_t \tag{6}$$

$$h_t = o_t \odot \tanh(C_t) \tag{7}$$

where $i_t$, $f_t$, and $o_t$ are the input, forget, and output gates at time $t$, $\tilde{C}_t$, $C_t$ and $h_t$ are the candidate cell state, the cell state and the output at time $t$, and $W$, $b$ are the weight matrix and bias vector, respectively.

XGBoost is a gradient boosting tree-based model which is also widely used in time series forecasting. It uses a strategy similar to gradient descent to iteratively train multiple decision trees, each of which is built to minimize the objective function. XGBoost performs well in dealing with high-dimensional sparse data and nonlinear features, which can be written as:

$$\hat{y}_i = \sum_{k=1}^{K} f_k(x_i) \tag{8}$$

where $\hat{y}_i$ is the prediction for the $i$-th observation $x_i$, $f_k$ is the $k$-th among the total $K$ decision trees. To obtain $f_k$, the algorithm aims to find a set of parameters $\theta_k$ by minimizing the following objective function:

$$\mathcal{L}(\theta_k) = \sum_{i=1}^{n} l(y_i, \hat{y}_i) + \sum_{k=1}^{K} \Omega(f_k) \tag{9}$$

where $l(y_i, \hat{y}_i)$ is the loss function, $\Omega(f_k)$ is the regularization term on the $k$-th tree, and $n$ is the number of observations in the training set. The algorithm then iteratively adds trees to the ensemble, using the gradient of the objective function with respect to the parameters to adjust the weights and learn the optimal values of $\theta$.

Ensemble is not a single ML algorithm, but by constructing and combining multiple base learners to complete learning tasks. Stacking is one of them, which inputs the predictions of several base learners into a metamodel as a new training set. This method can make full use of the advantages of the basic models, and reduce the deviation and variance to achieve better forecast performance. Based on this idea, in this paper, the core of ECLF is designed by incorporating LSTM as the error correction model to forecast trend and random components using the features selected in the previous step, and then inputs the forecast results as new features to feed into XGBoost for final forecasting of the two components, meanwhile, takes the values of same period last year of the seasonal component as its forecast results. Benefiting from this design, ECLF has the ability to capture the complex and nonlinear relationships of the two components and enhance the accuracy by incorporating previous forecast results as new features to reduce accidental factors. Therefore, by combining LSTM and XGBoost, the strengths of both models can be leveraged to achieve more accurate and robust LF in real-world applications.

### III. NUMERICAL EXPERIMENT

#### A. Basic Setup

*a) Datasets:* In this paper, the total electricity load consumption of two Chinese cities named D1 and D2 are adopted as datasets. These datasets cover monthly real load data and 12 features corresponding to load consumption, such as off days, average temperature, humidity, wind speed, rainfall, air pressure, cloudiness, and maximum and minimum temperature for each month from Jan. 2013 to Dec. 2021.

*b) Baselines:* ECLF is tested against eleven relevant state-of-the-art models in the field of electricity LF, which consist of two classical time series models, including X-12-ARIMA and TES, five ML models, including SVR [25], XGBoost [26], LSTM [27], general regression neural network (GRNN) [28] and CNN [29], and four advanced hybrid ML models, including APLF [30], NBEATS [31], TCN [32] and ETS+RD-LSTM [33].

*c) Evaluation Indicators:* Mean absolute error (MAE) and mean absolute percentage error (MAPE) are commonly used for evaluating the absolute error and the error ratio between the forecast results and the ground truths [34-38], which are calculated by the following equations:

$$MAE = \frac{1}{n}\sum_{i=1}^{n} |y_i - \hat{y}_i|, \quad MAPE = \frac{100\%}{n}\sum_{i=1}^{n} \left|\frac{y_i - \hat{y}_i}{y_i}\right| \tag{10}$$

where $y_i$ and $\hat{y}_i$ are respectively the observation and forecast for month $i$, and $n$ is the total number of months forecasted.

*d) Experimental Design:* The data from Jan. to Dec. 2021 on both datasets are adopted as the test set and the rest are adopted as the training set. The features selected for trend component include month, temperature and off days, while for random component include temperature and off days. All the hyperparameters of ECLF are selected and optimized on the training set, in which 50 units, batch_size of 12 and 120 epochs are adopted for LSTM while n_estimators of 300, max_depth of 2 and learning_rate of 0.11 are adopted for XGBoost. It is worth mentioning that the Adam optimizer is applied to the LSTM training process, for its momentum mechanism and ability to adjust the learning rate adaptively can accelerate the convergence and improve the computational efficiency [39-41]. The whole experiment is conducted on a laboratory computer with a 3.4-GHz i7 CPU and a 16-GB RAM.

#### B. Experimental Result

*a) Comparison Results of MAE:* The comparison results of MAE on the two datasets are shown in Fig. 2, from where the following observations can be seen.

- Almost all the models achieve higher accuracy in the latter half of the year than that in the first half. Especially, the forecast errors for Jan. to Feb. and Jul. to Aug. are relatively higher. This may be caused by uncatchable big swings in weather and holiday factors.

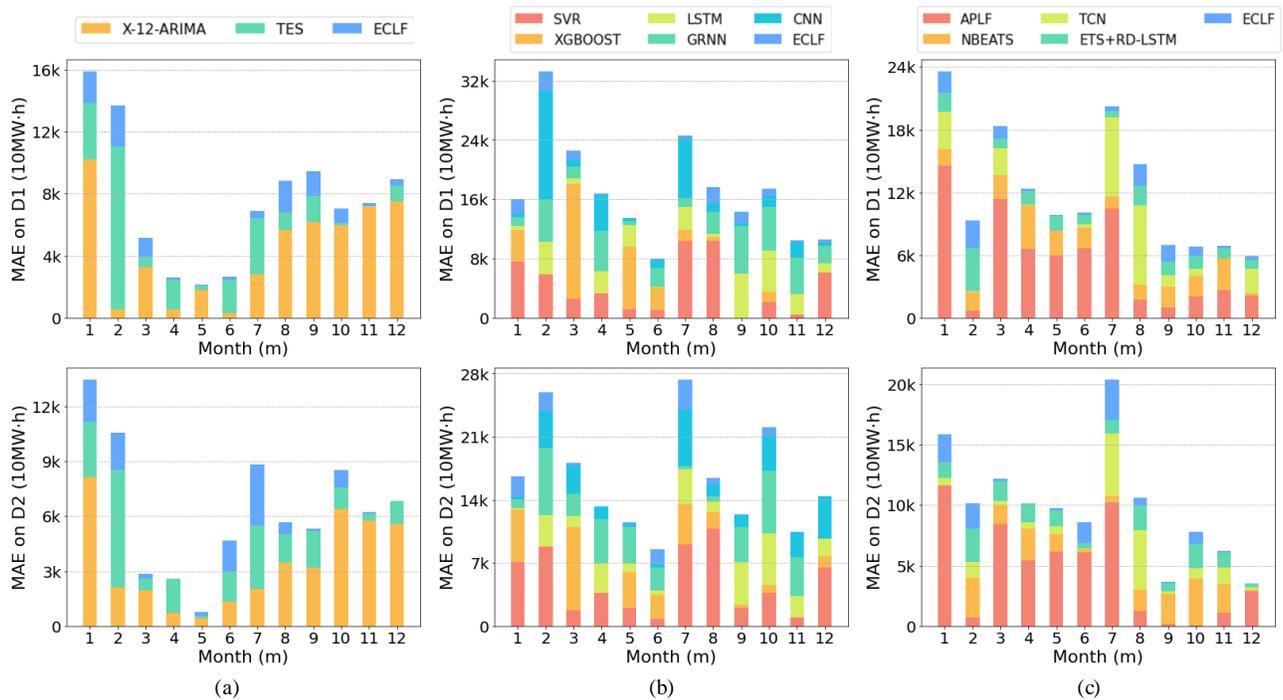

Fig. 2. Comparsion results of MAE with (a) classical time series models, (b) ML models and (c) advanced hybrid ML models.

- The classical time series model TES performs better than ML models. The errors of the ML models, such as CNN and XGBoost, fluctuate greatly, which means their accuracy is sometimes high and sometimes low.
- The hybrid models outperform both classical models and ML models except for APLF. ECLF outperforms all the other ML models, validating incorporating error correction modules can enhance single ML methods.
- Of all the models compared, ECLF performs the best, achieving the lowest mean-MAE over 12 months and the most consistent forecast results in most cases.

*b) Comparison Results of MAPE:* The comparison results of MAPE on D1 and D2 are recorded in Table I and Table II with some statistical analysis. To be specific, the mean-MAPEs over 12 months are recorded in the fourth row to the last, and the scores of win/loss are recorded in the third row to the last, where a score of win/loss of i/j for a baseline means its accuracy is higher on i-months and lower on j-months than ECLF. The comprehensive ranking is recorded in the penultimate row, where a smaller F-rank value indicates higher accuracy. The Wilcoxon signed-rank test is used to test whether the MAPE of ECLF was significantly lower than that of the baselines, and the accepted hypothesis with a significance level of 0.05 are shown in bold in the last row. The following conclusions can be drawn from the table.

- ECLF achieves lower MAPE and variance in most cases, with an average of 2.31% ±1.91% on D1 and 2.42% ±1.78% on D2, respectively. Specifically, ECLF loses 38 cases and wins 94 cases on D1, and 28 and 104 on D2, proving that ECLF has lower and more stable forecast errors than the other models.
- ECLF ranks ahead of all the others on both datasets, indicating that it achieves the highest forecast accuracy among the comparison models.
- Almost all the p-values are smaller than 0.05, proving that ECLF significantly outperforms each of the comparison models. Note that although the hypothesis is not accepted in six cases in total, the proposed model still has much lower mean-MAPE and variance than the others.

*c) Ablation Study:* The ablation experiment is designed to find out how the error correction mechanism affects the performance of ECLF. The results of MAPE are shown in Fig. 3, where the forecast error generated by the complete model and model without error correction (i.e. instead of using the trend and random components forecasted by LSTM as new features, use XGBoost directly) are labeled *ECLF* and *EC-Neither*, respectively, and the forecast error generated by error correction for random component only and for trend component only are labeled *EC-RC* and *EC-TC*, respectively. The following observations can be drawn from the figure.

- ECLF achieves the lowest MAPE, while EC-Neither shows its worst accuracy on both datasets, indicating that error correction is essential for both trend and random components to achieve better performance.

TABLE I. COMPARSION RESULTS OF MAPE (%) ON D1

| Month | X-12-ARIMA | TES | SVR | XGBOOST | LSTM | GRNN | CNN | APLF | NBEATS | TCN | ETS+RD-LSTM | ECLF |
|---|---|---|---|---|---|---|---|---|---|---|---|---|
| 1 | 20.02 | 7.05 | 14.84 | 8.40 | 1.00● | 2.23● | 0.95● | 28.46 | 3.08● | 6.99 | 3.59● | 3.99 |
| 2 | 1.92● | 35.19 | 19.53 | 0.11● | 14.93 | 18.81 | 49.48 | 2.30● | 6.15● | 0.24● | 13.69 | 8.79 |
| 3 | 6.88 | 1.43● | 5.39 | 32.80 | 1.69● | 3.38 | 1.86● | 24.04 | 4.76 | 5.51 | 1.87● | 2.58 |
| 4 | 1.24 | 4.37 | 7.57 | 0.01● | 7.07 | 12.40 | 11.26 | 15.21 | 9.64 | 0.11● | 3.11 | 0.39 |
| 5 | 4.13 | 0.63 | 2.53 | 19.42 | 6.67 | 1.21 | 0.95 | 13.71 | 5.37 | 0.09● | 3.28 | 0.14 |
| 6 | 0.68 | 4.69 | 2.40 | 6.45 | 0.29● | 5.60 | 2.29 | 14.55 | 4.22 | 0.70 | 2.00 | 0.44 |
| 7 | 5.15 | 6.59 | 18.81 | 2.89 | 5.54 | 2.26 | 14.59 | 19.04 | 2.14 | 13.80 | 1.07 | 0.81 |
| 8 | 10.14 | 2.07● | 18.70 | 0.91● | 0.76● | 5.28 | 2.28● | 3.06● | 2.69● | 13.62 | 3.40● | 3.66 |
| 9 | 13.87 | 3.73 | 0.05● | 0.01● | 13.34 | 14.28 | 0.77● | 2.13● | 4.47 | 2.52● | 2.95● | 3.54 |
| 10 | 14.19 | 0.34● | 5.01 | 3.30 | 13.36 | 13.75 | 3.72 | 4.88 | 4.66 | 1.60● | 2.97 | 2.15 |
| 11 | 16.47 | 0.04● | 0.96 | 0.00● | 6.27 | 11.41 | 4.76 | 5.96 | 7.06 | 0.05● | 2.20 | 0.46 |
| 12 | 15.05 | 2.05 | 12.09 | 0.21● | 2.43 | 4.65 | 0.95 | 4.25 | 0.40● | 4.68 | 1.71 | 0.75 |
| Mean-MAPE | 9.15±2.70 | 5.68±3.17 | 8.99±7.42 | 6.21±10.1 | 6.11±5.27 | 7.94±5.86 | 7.82±13.84 | 11.47±9.05 | 4.55±2.42 | 4.16±5.04 | 3.49±3.39 | 2.31±1.91 |
| Win/Loss | 1/11 | 4/8 | 1/11 | 6/6 | 4/8 | 1/11 | 4/8 | 3/9 | 4/8 | 6/6 | 4/8 | 94/38 |
| F-rank | 8.500 | 5.500 | 8.250 | 5.083 | 6.167 | 8.333 | 5.833 | 9.000 | 6.500 | 5.417 | 5.250 | 4.167 |
| P-value | **0.0046** | **0.0447** | **0.0034** | 0.2349 | **0.0171** | **0.0024** | 0.0881 | **0.0081** | **0.032** | 0.2349 | **0.0261** | - |

TABLE II. COMPARSION RESULTS OF MAPE (%) ON D2

| Month | X-12-ARIMA | TES | SVR | XGBOOST | LSTM | GRNN | CNN | APLF | NBEATS | TCN | ETS+RD-LSTM | ECLF |
|---|---|---|---|---|---|---|---|---|---|---|---|---|
| 1 | 18.25 | 6.84 | 15.99 | 13.01 | 0.28● | 2.23● | 0.64● | 26.01 | 0.18● | 1.32● | 2.98● | 5.10 |
| 2 | 6.95 | 21.20 | 29.13 | 0.21● | 11.43 | 24.52 | 13.47 | 2.26● | 11.02 | 4.34● | 9.04 | 6.82 |
| 3 | 4.79 | 1.68 | 4.29 | 22.95 | 2.95 | 6.08 | 8.01 | 20.91 | 3.85 | 0.95 | 3.96 | 0.55 |
| 4 | 2.03 | 5.06 | 10.04 | 0.01● | 8.95 | 13.58 | 3.68 | 14.87 | 7.14 | 1.43 | 4.33 | 0.05 |
| 5 | 1.18 | 0.25● | 5.45 | 10.59 | 2.63 | 10.90 | 0.67 | 16.53 | 3.85 | 1.73 | 3.51 | 0.62 |
| 6 | 3.40● | 4.10● | 1.98● | 6.75 | 1.26● | 6.49 | 0.77● | 15.34 | 0.71● | 0.15● | 1.09● | 4.31 |
| 7 | 4.21● | 7.11 | 18.73 | 9.22 | 7.87 | 0.72● | 12.90 | 21.10 | 1.06● | 10.71 | 2.26● | 6.94 |
| 8 | 6.96 | 3.12 | 21.66 | 3.74 | 2.33 | 1.17● | 2.72 | 2.45 | 3.59 | 9.85 | 4.06 | 1.32 |
| 9 | 7.97 | 4.88 | 4.93 | 0.99 | 11.74 | 9.63 | 3.17 | 0.41 | 6.29 | 0.50 | 1.51 | 0.31 |
| 10 | 18.22 | 3.35 | 10.51 | 2.49● | 16.36 | 19.62 | 11.00 | 0.14● | 10.95 | 2.44● | 5.91 | 2.75 |
| 11 | 15.86 | 1.01 | 2.64 | 0.01● | 6.51 | 11.88 | 7.38 | 3.06 | 6.48 | 3.85 | 3.42 | 0.33 |
| 12 | 12.90 | 2.91 | 15.05 | 2.90 | 4.40 | 0.24 | 10.74 | 6.62 | 0.40 | 0.38 | 0.82 | 0.00 |
| Mean-MAPE | 8.56±2.24 | 5.13±3.45 | 11.70±8.54 | 6.07±6.91 | 6.39±4.98 | 8.92±7.71 | 6.26±4.90 | 10.81±9.30 | 4.63±3.83 | 3.14±3.58 | 3.57±2.34 | 2.42±1.78 |
| Win/Loss | 2/10 | 2/10 | 1/11 | 4/8 | 2/10 | 3/9 | 2/10 | 2/10 | 3/9 | 4/8 | 3/9 | 104/28 |
| F-rank | 8.083 | 5.917 | 9.000 | 6.167 | 6.917 | 7.917 | 6.917 | 7.833 | 5.750 | 4.417 | 5.750 | 3.333 |
| P-value | **0.0061** | **0.0024** | **0.0007** | **0.032** | **0.0212** | **0.0134** | **0.0105** | **0.0061** | 0.0757 | 0.2593 | 0.1331 | - |

- EC-RC performs better than EC-TC. This may be because random sub-series forecasting is more in need of error correction for its randomness and complexity of influencing factors.

## IV. CONCLUSION

This paper proposes a hybrid model for mid-term electricity LF called ECLF, where STL is used to decompose the load data into seasonal and non-seasonal components, and stacking ensemble is employed to combine LSTM and XGBoost to construct error correction based forecasting modules, which results in improved LF accuracy. The experimental results verifies the effectiveness of ECLF, showing that ECLF significantly outperforms its peers in terms of mid-term LF stability and accuracy. In future research, some intelligent optimization algorithms such as latent factor analysis [42-46] will be introduced into the model to handle the missing data issue of data processing and feature selection [47-50], thus improving the performance while dealing with LF tasks with different time scales.

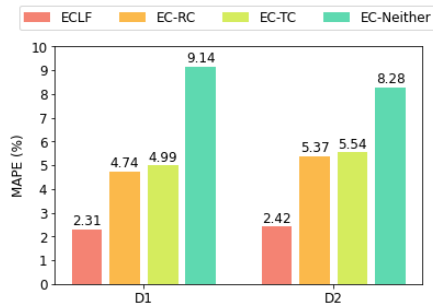

Fig. 3. Results of ablation study.